\documentclass[11pt]{article}
\usepackage{geometry}
\usepackage{coling2020}
\usepackage{times}
\usepackage{url}
\usepackage{latexsym}
\usepackage{microtype}
\usepackage[T2A,T1]{fontenc}
\usepackage[utf8]{inputenc}
\usepackage{booktabs}
\usepackage{multirow}
\usepackage{graphicx}
\usepackage{rotating}
\usepackage{todonotes}
\usepackage{amsmath,amssymb}
\usepackage{siunitx}
\usepackage{xcolor}

\colingfinalcopy 

\title{UiO-UvA at SemEval-2020 Task 1:\\ Contextualised Embeddings for Lexical Semantic Change Detection}

\author{Andrey Kutuzov\\
  University of Oslo, Norway \\
  {\tt andreku@ifi.uio.no} \\\And
  Mario Giulianelli \\
  University of Amsterdam, Netherlands \\
  {\tt m.giulianelli@uva.nl} \\}

\date{}

\begin{document}
\maketitle
\begin{abstract}
We apply contextualised word embeddings to lexical semantic change detection in the SemEval-2020 Shared Task 1. This paper focuses on Subtask 2, ranking words by the degree of their semantic drift over time. We analyse the performance of two contextualising architectures (BERT and ELMo) and three change detection algorithms. We find that the most effective algorithms rely on the cosine similarity between averaged token embeddings and the pairwise distances between token embeddings. They outperform strong baselines by a large margin (in the post-evaluation phase, we have the best Subtask 2 submission for SemEval-2020 Task 1), but interestingly, the choice of a particular algorithm depends on the distribution of gold scores in the test set.
\end{abstract}

\section{Introduction}
\label{intro}

%
%
\blfootnote{
    %
    \hspace{-0.65cm}  
    This work is licensed under a Creative Commons 
    Attribution 4.0 International Licence.
    Licence details:
    \url{http://creativecommons.org/licenses/by/4.0/}.
}

Words change their meaning over time in all natural languages. The emergence of large representative historical corpora and powerful data-driven semantic models has allowed researchers to track meaning change.
SemEval-2020 Shared Task 1 challenges its participants to classify a list of target words into stable or changed (Subtask 1) and/or to rank these words by the degree of their semantic change (Subtask 2) \cite{schlechtweg2020semeval}. The task is multilingual: it includes four lists of target words, respectively for English, German, Latin, and Swedish (several dozen words each). Each word list is accompanied with two historical corpora of varying size, consisting of texts created in two different time periods. The shared task organisers additionally provided frequency-based and distributional baseline methods.

We participated in Subtask 2 as the \textbf{UiO-UvA }team
and investigated the potential of \textit{contextualised embeddings} for the detection of lexical semantic change. We did not specifically focus on the binary Subtask 1 and used a different method there, which we do not report in this paper. Our systems are based on ELMo \cite{peters2018} and BERT \cite{devlin2019} language models, and employ 3 different algorithms to compare contextualised embeddings diachronically. Our \textit{evaluation phase} submission to the shared task ranked 10\textsuperscript{th} out of 34 participating teams, while in the \textit{post-evaluation phase}, our submission is the best (but some knowledge of the test sets statistics was needed, see below).
In this paper, we extensively evaluate all combinations of architectures, training corpora and change detection algorithms, using 5 test sets in 4 languages. 
 
Our main findings are twofold: 1) In 3 out of 5 test sets, ELMo consistently outperforms BERT, while being much faster in training and inference; 2) Cosine similarity of averaged contextualised embeddings and average pairwise distance between contextualised embeddings are the two best-performing change detection algorithms, but different test sets show strong preference to either the former or the latter. This preference shows strong correlation with the distribution of gold scores in a test set. While it may indicate that there is a bias in the available test sets, this finding remains yet unexplained.

Our implementations of all the evaluated algorithms are available at \url{https://github.com/akutuzov/semeval2020}, and the ELMo models we trained can be downloaded from the NLPL vector repository at \url{http://vectors.nlpl.eu/repository/} \cite{nlpl:2017}.

\section{Background}
Lexical semantic change detection (LSCD) task is determining whether and/or to what extent the meaning of a set of target words has changed over time, with the help of time-annotated corpora \cite{kutuzov:survey,tahmasebi2018survey}. 
LSCD is often addressed using distributional semantic models: time-sensitive word representations (so-called `diachronic word embeddings') are learned and then compared between different time periods.
This is a fast developing NLP sub-field with a number of influential papers (\cite{baroni:2011,kim2014,kulkarni2015,hamilton2016diachronic,bamler2017dynamic,kutuzov-etal-2017-temporal,deltredici2019short,Dubossarskyetal19}, and many others).

Most of the previous work used some variation of `static' word embeddings where each occurrence of a word form is assigned the same vector representation independently of its context. Recent contextualised architectures allow for overcoming this limitation by taking sentential context into account when inferring word token representations. However, application of such architectures to diachronic semantic change detection was up to now limited \cite{hu2019diachronic,giulianelli2020,addmore_2020}. While all these studies use BERT as their contextualising architecture, we extend our analysis to ELMo and perform a systematic evaluation of various semantic change detection approaches for both language models.

ELMo \cite{peters2018} was arguably the first contextualised word embedding architecture to attract wide attention of the NLP community. The network architecture consists of a two-layer Bidirectional LSTM on top of a convolutional layer.
BERT \cite{devlin2019} is a Transformer with self-attention trained on masked language modelling and next sentence prediction.
While Transformer architectures have been shown to outperform recurrent ones in many NLP tasks, ELMo allows faster training and inference than BERT, making it more convenient to experiment with different training corpora and hyperparameters. 

\section{System overview}
\label{section:system-overview}
Our approach relies on contextualised representations of word occurrences often referred to as \textit{contextualised embeddings)}. 
Given two time periods $t_1, t_2$ , two corpora $C_1, C_2$, and a set of target words, we use a neural language model to obtain \textit{contextualised embeddings} of each occurrence of the target words in $C_1$ and $C_2$ and use them to compute a continuous change score. This score indicates the degree of semantic change undergone by a word between $t_1$ and $t_2$, and the target words are ranked by its value. 

More precisely, given a target word $w$ and its sentential context $s = (v_1, ..., v_i, ..., v_m)$ with $w = v_i$, we extract the activations of a language model's hidden layers for sentence position $i$. The $N_w$ contextualised embeddings collected for $w$ can be represented as the usage matrix $\textbf{U}_w = \left( \textbf{w}_1, \ldots, \textbf{w}_{N_w} \right)$. The time-specific usage matrices $\textbf{U}_w^1,  \textbf{U}_w^2$ for time periods $t_1$ and $t_2$ are used as input to all the tested metrics of semantic change. 
We use three change detection algorithms:

\paragraph{1. Inverted cosine similarity over word prototypes (PRT)}
Given two usage matrices $\textbf{U}_w^{t_1}, \textbf{U}_w^{t_2}$, the degree of change of $w$ is calculated as the inverted similarity
\footnote{We also tried to use cosine distance ($1-\mathop{d} $) instead of inverted cosine similarity, but the results were marginally worse.} between the average token embeddings (`prototypes') of all occurrences of $w$ in the two time periods:
\begin{align}
    PRT\left(\textbf{U}_w^{t_1}, \textbf{U}_w^{t_2}\right) = \frac{1}{d\left(\frac{\sum_{\textbf{x}_i \in \textbf{U}_w^{t_1}} \textbf{x}_i}{N_w^{t_1}}, \frac{\sum_{\textbf{x}_j \in \textbf{U}_w^{t_2}} \textbf{x}_j}{N_w^{t_2}}\right)}  
\end{align}
where $N_w^{t_1}$ and $N_w^{t_2}$ are the number of occurrences of $w$ in time periods $t_1$ and $t_2$, and $d$ is a similarity metric, for which we use cosine similarity. 
This method is similar to the standard LSCD workflow with \textit{static} embeddings produced by Procrustes-aligned time-specific distributional models \cite{hamilton2016diachronic}, with the only additional step of averaging token embeddings to create a single vector. Since we want the algorithm to produce higher scores for the words which changed more, the inverted value of cosine similarity is used as the prediction.

\paragraph{2. Average pairwise cosine distance between token embeddings (APD)}
Here, the degree of change of $w$ is measured as the average distance between any two embeddings from different time periods \cite{giulianelli2020}: 
\begin{align}
    APD\left(\textbf{U}_w^{t_1}, \textbf{U}_w^{t_2}\right) =  \frac{1}{N_w^{t_1} \cdot N_w^{t_2}} \sum_{\textbf{x}_i \in \textbf{U}_w^{t_1},\ \textbf{x}_j \in \textbf{U}_w^{t_2}} d\left(\textbf{x}_i, \textbf{x}_j\right)
\end{align}
where $d$ is the cosine distance. High APD values indicate stronger semantic change.

\paragraph{3. Jensen-Shannon divergence (JSD)}
This measure relies on the partitioning of embeddings into clusters of similar word usages. We follow \newcite{giulianelli2020} and create a single usage matrix with occurrences from two corpora $[\textbf{U}^{t_1}_w ; \textbf{U}^{t_2}_w]$. We then standardise it and cluster its entries using Affinity Propagation \cite{frey2007affinity}, which automatically selects a number of clusters for each word \cite{addmore_2020}.
Finally, we define probability distributions $\textbf{u}_w^{t_1}, \textbf{u}_w^{t_2}$ based on the normalised counts of word occurrences from each cluster \cite{hu2019diachronic} and compute a JSD score \cite{lin1991jsd}:
\begin{align}
\begin{split}
\operatorname{JSD}(\textbf{u}_w^{t_1}, \textbf{u}_w^{t_2}) &= \operatorname{H}\left(\frac{1}{2} \left( \textbf{u}_w^{t_1} + \textbf{u}_w^{t_2}\right) \right) - \frac{1}{2} \left(\operatorname{H}\left(\textbf{u}_w^{t_1}\right) - \operatorname{H}\left(\textbf{u}_w^{t_2}\right)\right)
\end{split}
\end{align}
JSD scores measure the amount of change in the proportions of word usage clusters across time periods. A high JSD score indicates a high degree of lexical semantic change.

\section{Experimental setup} \label{sec:experimental}
For each of the 4 languages of the shared task, we train 4 ELMo model variants: 1) \textbf{Pre-trained}, an ELMo model trained on the respective Wikipedia corpus (English, German, Latin or Swedish)\footnote{The Wikipedia corpora were lemmatised using UDPipe \cite{udpipe:2017} prior to training.};
2) \textbf{Fine-tuned}, the same as Pre-trained but further fine-tuned on the union of the two test corpora; 3) \textbf{Trained on test}, trained only on the union of the two test corpora; 4) \textbf{Incremental}, two models---the first is trained on the first test corpus, and the second is the same model further trained on the second test corpus. 
The ELMo models are trained for 3 epochs (except English and Latin \textbf{Trained on test} and \textbf{Incremental} models, for which we use 5 epochs, due to small test corpora sizes), with the LSTM dimensionality of 2048, batch size 192 and 4096 negative samples per batch. All the other hyperparameters are left at their default values.\footnote{To train and fine-tune ELMo models, we use the code from \url{https://github.com/ltgoslo/simple_elmo_training}, which is essentially the reference ELMo implementation updated to the recent TensorFlow versions.}

For BERT, we use the \emph{base} version, with 12 layers and 768 hidden dimensions.\footnote{We rely on \textit{Hugging Face}'s implementation of BERT (available at \url{https://github.com/huggingface/transformers}, version 2.5.0), and follow their model naming conventions: \url{https://huggingface.co/models}.} For English, German and Swedish, we employ language-specific models: \textit{bert-base-uncased}, \textit{bert-base-german-cased}, and \textit{af-ai-center/bert-base-swedish-uncased}. For Latin, we resort to \textit{bert-base-multilingual-cased}, since there is no specific Latin BERT available yet. 
Given the limited size of the test corpora (in the order of $10^8$ word tokens at max), we do not train BERT from scratch and only test the \textbf{Pre-trained} and \textbf{Fine-tuned} BERT variants. The fine-tuning is done with BERT's standard objective for 2 epochs (English was trained for 5 epochs).
We configure BERT's WordPiece tokeniser to never split any occurrences of the target words (some target words are split by default into character sequences) and we add unknown target words to BERT's vocabulary. We perform this step both before fine-tuning and before the extraction of contextualised representations.

At inference time, we use all ELMo and BERT variants to produce contextualised representations of all the occurrences of each target word in the test corpora. For the \textbf{Incremental} variant, the representations for the occurrences in each of the two test corpora are produced using the respective model trained on this corpus. The resulting embeddings are of size $12 \times 768$ and $3 \times 512$ for BERT and ELMo, respectively. We employ three strategies to reduce their dimensionality to that of a single layer: 1) using only the top layer, 2) averaging all layers, 3) averaging the last four layers (only for BERT embeddings, as this aggregation method was shown to work on par with the all-layers alternative in \cite{devlin2019}).
Finally, to predict the strength of semantic change of each target word between the two test corpora, 
we feed the word's contextualised embeddings into the three algorithms of semantic change estimation described in Section \ref{section:system-overview}. We then compute the Spearman correlation of the estimated change scores with the gold answers. This is the evaluation metric of Subtask 2, and we use it throughout our experiments.

\section{Results} \label{sec:results}

\paragraph{Our submission.} In our `official' shared task submission in the evaluation phase we used top-layer ELMo embeddings with the cosine similarity change detection algorithm for all languages. English and German ELMo models were trained on the respective Wikipedia corpora. For Swedish and Latin, pre-trained ELMo models were not available, so we trained our own models on the union of the test corpora. This combination of architectures and algorithms was chosen based on our preliminary experiments with the available human-annotated semantic change datasets for English \cite{baroni:2011}, German \cite{schlechtweg2018durel} and Russian \cite{kutuzov2019russiandia}. 
The resulting Spearman correlations were 0.136 for English, 0.695 for German, 0.370 for Latin, and 0.278 for Swedish.
With an average Codalab score of 0.37, this submission ranked 10\textsuperscript{th} out of 34 teams in the evaluation phase. 

We were aware that the submitted setup was likely sub-optimal as it did not include the \textbf{Fine-tuned} model variant. After the official submission deadline, we finished training and fine-tuning all of our language models. Their systematic evaluation is the main contribution of this paper.

\begin{table}[ht]
    \centering
    \begin{tabular}{lllll}
    \toprule
    \multirow{2}{*}{\textbf{Baselines}}&  \multicolumn{2}{l}{\textbf{Frequency (FD)}}  & & -0.083 \\
    &    \multicolumn{2}{l}{\textbf{Count (CNT+CI+CD)}}  & & 0.144*  \\
    \midrule
    \multirow{2}{*}{\textbf{CBOW cosine distance}} & \multicolumn{2}{l}{\textbf{Incremental}}  & & 0.140 \\
    &  \multicolumn{2}{l}{\textbf{Procrustes}}  & & 0.392*** \\
    \midrule
    \multicolumn{2}{l}{\textbf{Contextualised embeddings}} & {\textbf{Top layer}} & \textbf{Average all layers} & \textbf{Average top 4 layers} \\
    \midrule
     & \multicolumn{4}{c}{\textbf{Cosine similarity (PRT)}} \\
    \multirow{2}{*}{\textbf{BERT}} & Pre-trained & 0.278** & 0.233 & 0.229 \\
                  & Fine-tuned & 0.373** & 0.320** & 0.338**\\
                  \cmidrule(lr){2-5}
\multirow{4}{*}{\textbf{ELMo}} & Pre-trained & 0.375** & 0.344** & {--} \\
 & Fine-tuned & 0.402** & 0.389** & {--} \\
 & Trained on test & 0.370** & 0.342** & {--} \\
 & Incremental & 0.114* & 0.127 & {--} \\
    \midrule
     & \multicolumn{4}{c}{\textbf{Pairwise distance (APD)}} \\
    \multirow{2}{*}{\textbf{BERT}} & Pre-trained & 0.237** & 0.163* & 0.203*\\
 & Fine-tuned & 0.363*** & 0.241** & 0.297* \\
 \cmidrule(lr){2-5}
    \multirow{4}{*}{\textbf{ELMo}} & Pre-trained & 0.296** & 0.172* & {--}\\
 & Fine-tuned & 0.405*** & \textbf{0.406}*** &{--} \\
 & Trained on test & 0.338** & 0.295*** & {--}\\
 & Incremental & 0.126** & \hspace{-0.3em}-0.001* & {--}\\
    \midrule
    & \multicolumn{4}{c}{\textbf{Jensen-Shannon divergence (JSD)}} \\
    \multirow{2}{*}{\textbf{BERT}} & Pre-trained & 0.181* & 0.125 & 0.203* \\
 & Fine-tuned & 0.176* & 0.223** & 0.186** \\
 \cmidrule(lr){2-5}
    \multirow{4}{*}{\textbf{ELMo}} & Pre-trained & 0.251* & 0.196* & {--}\\
& Fine-tuned & 0.197* & 0.156* &{--} \\
& Trained on test & 0.225* & 0.163* & {--}\\
& Incremental & \hspace{-0.3em}-0.037 & \hspace{-0.3em}-0.009 & {--}\\
    \bottomrule
    \end{tabular}
    \caption{Spearman correlation coefficients for Subtask 2 averaged over four languages. The number of asterisks denotes the number of languages for which the correlation was statistically significant ($p < 0.05$).}
    \label{tab:test_task2_all}
\end{table}

\paragraph{Current results.} The average scores of all the tested configurations across 4 languages are given in Table \ref{tab:test_task2_all}. This table includes both the results of the configurations we used in the evaluation phase and the results of the configurations we tested after the submission deadline. We compare our scores to the organisers' baselines (FD and CNT+CI+CD, as provided by \newcite{schlechtweg2020semeval})
and the classical approach of calculating cosine distance between static CBOW word embeddings \cite{mikolov2013efficient}. The CBOW models were used in two different flavors: 1) `incremental', where the $C_2$ model was initialised with the $C_1$ weights \cite{kim2014}, and 2) `Procrustes', where the two models were trained independently on $C_1$ and $C_2$, and then aligned using the orthogonal Procrustes transformation \cite{hamilton2016diachronic}. Note that although we did not try different hyperparameter combinations for the static embeddings (varying vector dimensionalities, learning rates, vocabulary sizes, etc), we did not do that for contextualised embeddings either. All the training hyperparameters for both ELMo and BERT were fixed to their default values (see section~\ref{sec:experimental}), we varied only the training corpora and the layers from which embeddings were extracted. Thus, we believe that static and contextualised embedding-based approaches were under a fair comparison.

Table \ref{tab:test_task2_all} shows that no method achieves statistically significant correlation on \textit{all 4} languages, which attests both to the difficulty of the task and the diversity of the test sets. CBOW Procrustes is a surprisingly strong approach, consistently outperforming the organisers' baselines. Only PRT and APD obtain higher average scores, with fine-tuned ELMo models performing better than fine-tuned BERT. 

Judging only from the average correlation scores, contextualised embeddings do not seem to outshine their static counterparts, especially considering that both ELMo and BERT are more computationally demanding than CBOW. However, closer analysis of per-language results shows that in fact the contextualised approaches outperform the CBOW Procrustes baseline by a large margin for \textit{each} of the shared task test sets. Table \ref{tab:test_task2_best} features the scores obtained by our best-performing methods (PRT and APD with top layer embeddings from fine-tuned ELMo and BERT) on the individual languages of the shared task. We also report performance on the GEMS (`GEometrical Models of Natural Language Semantics workshop') test set \cite{baroni:2011} to enable a comparison with previous work \cite{giulianelli2020}.
The discrepancy between the averaged and the per-language results can be explained by properties of the test sets: APD works best on the English and Swedish sets, while PRT yields the best scores for German and Latin. 

Although consistency across languages (3 out of 4) is an important benefit of the CBOW Procrustes approach, with the right choice of APD or PRT, contextualised embeddings can improve Spearman correlation coefficients by up to 50\%. This is not a language-specific property: the English GEMS test set does not behave like the English test set from the shared task. In fact, one can observe 3 groups of test sets with regards to their statistical properties and to the method they favour: group 1 (Latin and German) exhibits rather uniform gold score distributions and  prefers PRT;  group 2 (English and Swedish) is characterised by more skewed gold score distributions and prefers APD; group 3 (GEMS) is in between, with no clear preference.

Interestingly, the method which produces a more uniform predicted score distribution (APD) works better for the test sets with skewed gold distributions, and the method which produces a more skewed predicted score distribution (PRT) works better for the uniformly distributed test sets (as can be seen in the Appendix). Furthermore, there is perfect negative correlation ($\rho=-1$) between the median gold score of a test set and the performance of the APD algorithm with fine-tuned ELMo models on this test set. The same correlation for the APD performance is not significant but strictly negative. We currently do not have a plausible explanation for this behaviour.

Table \ref{tab:test_task2_best} also supports the previous observation that ELMo models perform better than BERT in the LSCD task. The only test set for which this is not the case is Latin,\footnote{The Latin test corpora are very peculiar: 1) homonyms in them are followed by `\#' and the sense identifier, which is not the case for Latin Wikipedia, 2) the sizes of $C_1$ and $C_2$ are very imbalanced, with the latter being 4 times larger than the former.}
while on GEMS, ELMo and BERT are on par.\footnote{Note that \cite{addmore_2020} report a Spearman correlation of 0.510 on the GEMS dataset using fine-tuned BERT embeddings with Affinity Propagation and JSD. However, we were unable to reproduce these results, even when using the published code.} One possible explanation is that our ELMo models were pre-trained on lemmatised Wikipedia corpora and thus better fit the test corpora, provided in lemmatised form by the organisers. The BERT models were pre-trained on raw corpora, and fine-tuning them on lemmatised data proves less successful. 
This is of course not an advantage of the ELMo architecture \textit{per se}; however, easy and fast training from scratch on the respective Wikipedia corpora for each shared task language was possible only because of much lower computational requirements of ELMo compared to BERT.

\begin{table}
    \centering
    \begin{tabular}{ll|lllll}
&\textit{Algorithm} & \textit{English} & \textit{German} & \textit{Latin} & \textit{Swedish} & \textit{GEMS} \\
\midrule
\multirow{2}{*}{\textit{CBOW}} & Incremental & 0.210 & 0.145 & 0.217 & -0.012 & \textbf{0.424}$^{\dagger}$  \\
&Procrustes & 0.285 & 0.439$^{\dagger}$ & 0.387$^{\dagger}$ & 0.458$^{\dagger}$ & 0.235$^{\dagger}$  \\
\midrule
&\multicolumn{6}{c}{\textbf{Fine-tuned contextualised embeddings (top layer)}}  \\
\midrule
\multirow{2}{*}{\textit{ELMo}} & Cosine similarity (PRT) & 0.254 & \textbf{0.740}$^{\dagger}$ & 0.360$^{\dagger}$ & 0.252 & 0.323$^{\dagger}$  \\
& Average pairwise distance (APD) & \textbf{0.605}$^{\dagger}$ & 0.560$^{\dagger}$ & -0.113 & \textbf{0.569}$^{\dagger}$ & 0.323$^{\dagger}$  \\
\midrule
\multirow{2}{*}{\textit{BERT}} & Cosine similarity (PRT) & 0.225 & 0.590$^{\dagger}$ & \textbf{0.561}$^{\dagger}$ & 0.185 & 0.394$^{\dagger}$  \\
& Average pairwise distance (APD) & 0.546$^{\dagger}$ & 0.427$^{\dagger}$ & 0.372$^{\dagger}$ & 0.254 & 0.243$^{\dagger}$  \\
    \end{tabular}
    \caption{Spearman correlation per test set for our best methods. $\dagger$ marks statistical significance ($p < 0.05$).} 
    \label{tab:test_task2_best}
\end{table}

In the post-evaluation phase of the shared task, we submitted predictions obtained with the optimal system configurations: fine-tuned ELMo + APD for English and Swedish, fine-tuned ELMo + PRT for German, and fine-tuned BERT + PRT for Latin. It reached the average Spearman correlation of 0.618 and, at the time of writing, it is the best Subtask 2 submission for SemEval-2020 Task 1. Of course, the optimal choice of configurations was possible only because we already knew the test data. Still, it is useful for understanding of the real abilities of contextualised embedding-based approaches and the peculiarities of different models and test sets.

\section{Conclusion}
Our experiments for the SemEval-2020 Shared Task 1 (Subtask 2) show that using contextualised embeddings to rank words by the degree of their semantic change yields strong correlation with human judgements, outperforming static embeddings. Models pre-trained on large external corpora and fine-tuned on the historical test corpora produce the highest correlation results, with ELMo slightly but consistently outperforming BERT as a contextualiser.

Inverted cosine similarity between averaged contextualised embeddings and the average pairwise cosine distance between contextualised embeddings turned out to be the best semantic change detection algorithms. An interesting finding is that the former algorithm favours the test sets with uniform gold score distribution, while the latter works best with the test sets where the gold score distribution is skewed towards low values. This distinction is not related to the language of the test set, or any other linguistic or statistical property of the test sets we looked at. While slightly different human annotation protocols can be in play here, we believe this dependency between the gold scores distribution and the performance of semantic change detection systems deserves to be investigated further in future work.

We also intend to explore the possibilities to improve our best-performing methods (PRT and APD), especially with regards to removing the outlier embeddings before calculating the semantic change score.

\bibliographystyle{coling}
\bibliography{semeval2020}

\section*{Appendix A. Score distributions}

\begin{figure}[ht]
    \centering
    \includegraphics[scale=0.4,keepaspectratio]{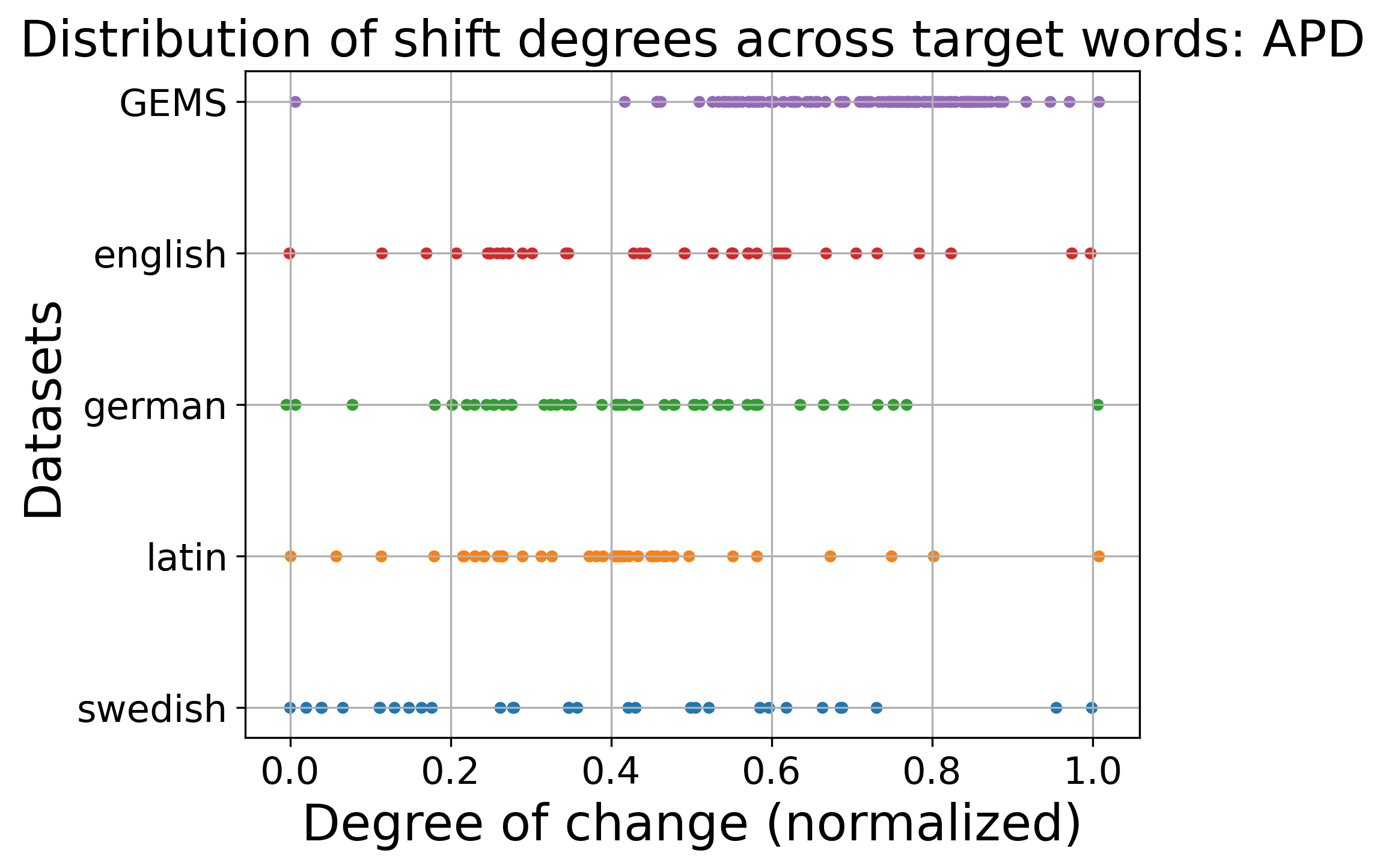}
    \includegraphics[scale=0.4,keepaspectratio]{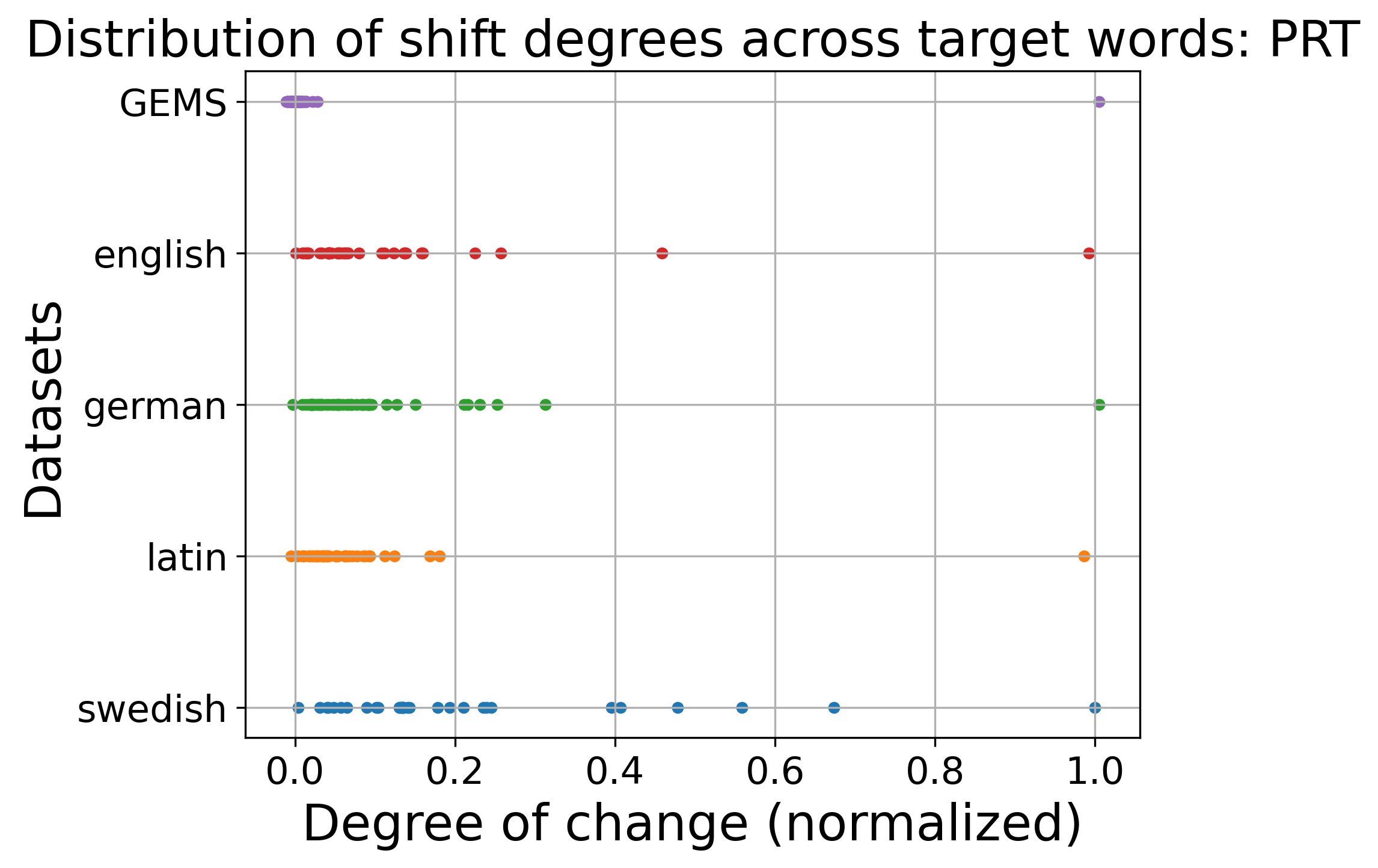}
    \includegraphics[scale=0.4,keepaspectratio]{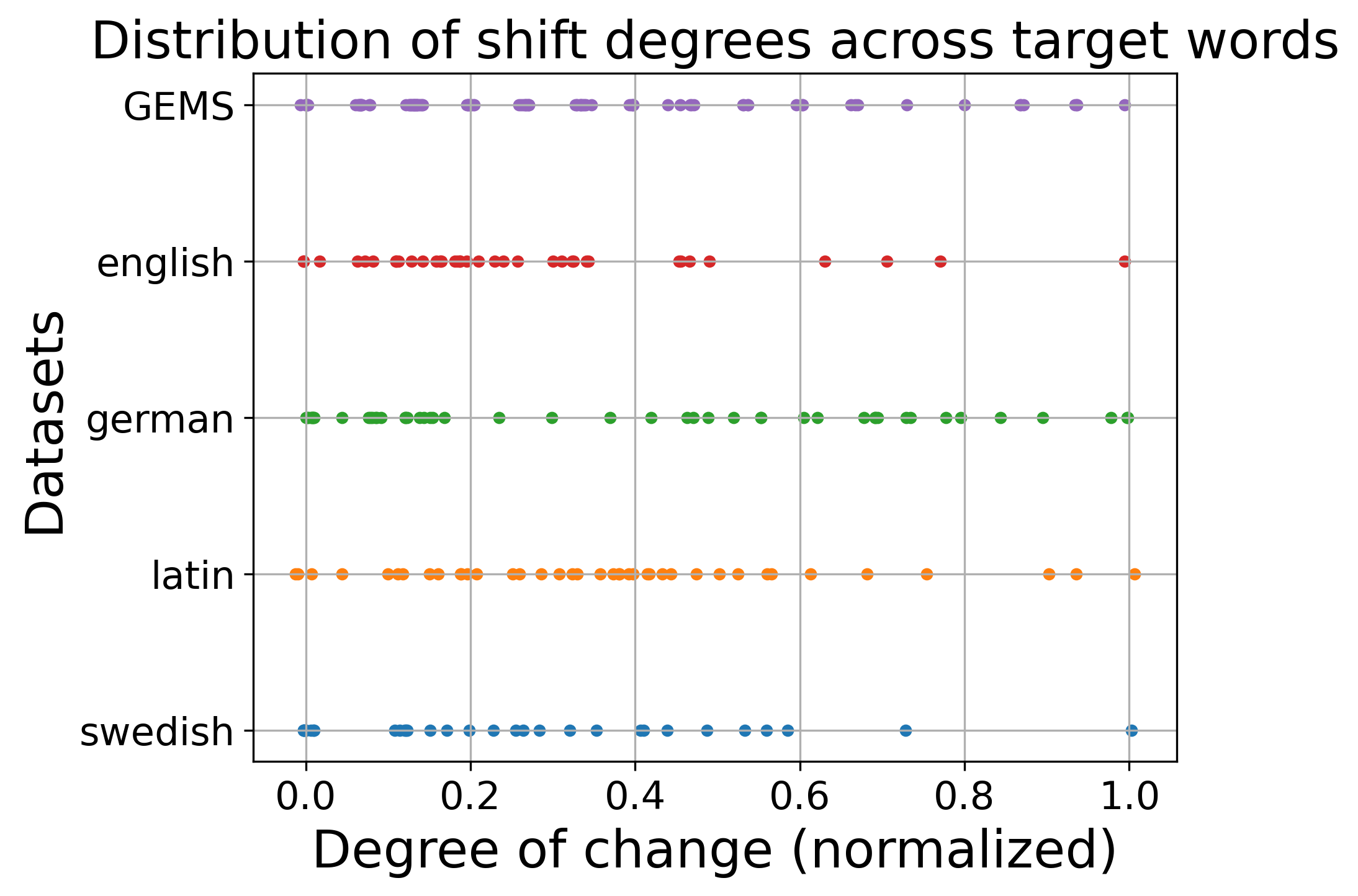}
    \caption{Bottom: distribution of semantic change degree in the \textbf{gold} data; top left: distribution of scores predicted by the \textbf{APD} algorithm; top right: distribution of scores predicted by the \textbf{PRT} algorithm.}
    \label{fig:shift_distrib}
\end{figure}

In the bottom part of Figure \ref{fig:shift_distrib}, we show how different the 5 test sets are in terms of how the \textit{gold} scores are distributed in them. In some test sets, the gold scores are skewed to the left, while some have a more uniform distribution. The top part of Figure \ref{fig:shift_distrib} shows the distributions of the \textit{predicted} scores produced by the APD and PRT algorithms (with fine-tuned ELMo embeddings). PRT tends to squeeze the majority of predictions near the lower boundary (no semantic change), with a low median score. In contrast, APD distributes its predictions in a much more uniform way, with a higher median score. Counter-intuitively, skewed gold distributions favour uniform predictions and vice versa.

\begin{figure}[ht]
    \centering
    \includegraphics[scale=0.8,keepaspectratio]{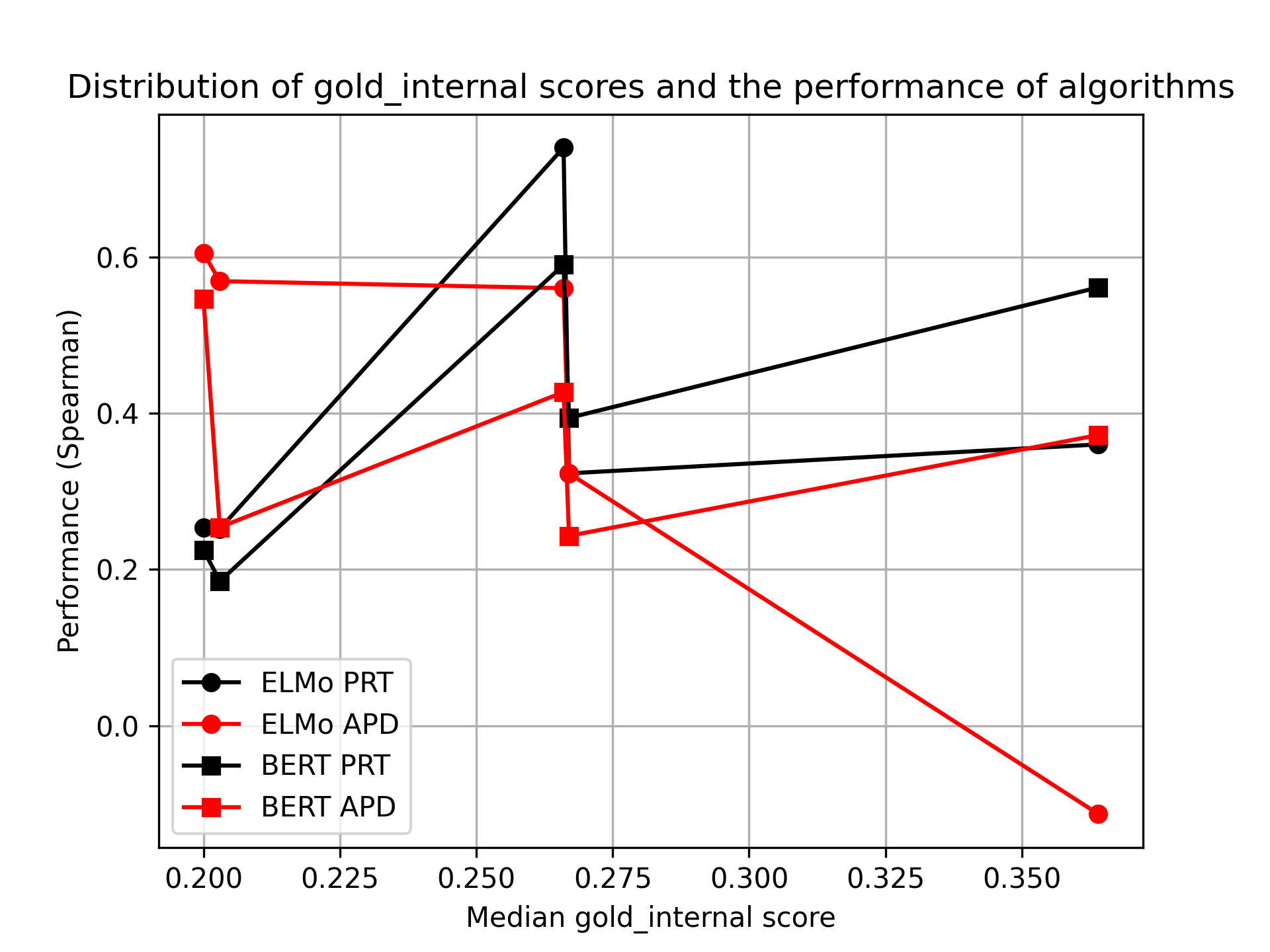}
    \caption{Performance of the PRT and APD algorithms depending on the median gold score.}
    \label{fig:performance}
\end{figure}

The grouping differences can be quantified with respect to the median gold score (after unit-normalisation). Figure \ref{fig:performance} shows the dependency of the PRT and APD performance on the median score of the gold test set. The dots here are the performance values of PRT or APD algorithms on different test sets. English and Swedish test sets are in the left part of the plot with the median gold scores of 0.200 and 0.203 correspondingly.  German, GEMS and Latin are on the right with 0.266, 0.267 and 0.364 correspondingly. There is a perfect negative Spearman correlation between the median gold scores of these 5 test sets and the performance of APD semantic change detection algorithm on each of them (with fine-tuned ELMo embeddings).

\end{document}